\documentclass[final,5p,times,authoryear,twocolumn]{elsarticle}
\usepackage[colorlinks, %%改变引用颜色
            linkcolor=blue,  
            anchorcolor=blue, 
            citecolor=blue,
            ]{hyperref}
\usepackage{graphicx}
\usepackage{subfigure}
\journal{NEURAL NETWORKS}
\usepackage{booktabs}
\usepackage{caption}
\usepackage{comment}
\usepackage{caption}
\usepackage{titlesec}
\titleformat{\subsection}
{\normalfont\bfseries}{\textbf{\thesubsection}}{1em}{\textbf}
\titleformat{\subsubsection}
{\normalfont\bfseries}{\textbf{\thesubsubsection}}{1em}{\textbf}
\usepackage{float}
\usepackage{afterpage}
\usepackage{amsmath}
\usepackage[T1]{fontenc}

\begin{document}
\begin{frontmatter}

\title{Lightweight texture transfer based on texture feature preset}
\author{ShiQi Jiang,JunJie Kang, YuJian Li\affiliation{organization={Guilin University of Electronic Technology},
            city={Guilin},
            country={China}}}

\begin{abstract}
In the task of texture transfer, reference texture images typically exhibit highly repetitive texture features, and the texture transfer results from different content images under the same style also share remarkably similar texture patterns. Encoding such highly similar texture features often requires deep layers and a large number of channels, making it is also the main source of the entire model's parameter count and computational load, and inference time. We propose a lightweight texture transfer based on texture feature preset (\textbf{TFP}). TFP takes full advantage of the high repetitiveness of texture features by providing preset universal texture feature maps for a given style. These preset feature maps can be fused and decoded directly with shallow color transfer feature maps of any content to generate texture transfer results, thereby avoiding redundant texture information from being encoded repeatedly. The texture feature map we preset is encoded through noise input images with consistent distribution (standard normal distribution). This consistent input distribution can completely avoid the problem of texture transfer differentiation, and by randomly sampling different noise inputs, we can obtain different texture features and texture transfer results under the same reference style. Compared to state-of-the-art techniques, our TFP not only produces visually superior results but also reduces the model size by 3.2-3538 times and speeds up the process by 1.8-5.6 times.
\end{abstract}


\end{frontmatter}

\section{Introduction}
Style transfer is a highly attractive image processing technique that can transfer the unique colors and texture styles of artworks to content images. In recent years, methods for style transfer have been widely proposed, which can be roughly divided into two categories: online image optimization and model optimization.

The representative of image optimization methods is (\cite{gatys}), which innovatively transfers gradients to the input image and iteratively optimizes the input content image directly. The style pattern is represented by the feature correlation of deep convolutional neural networks (VGG, \cite{vgg}). Subsequent work mainly focuses on different forms of loss functions (\cite{15,27}). However, this slow online optimization method has a high time cost and greatly reduces its actual citation value. In contrast, the model optimization method effectively solves the time-consuming problem of online iteration through offline model training and forward reasoning. There are three main types of model optimization: (1) Training exclusive style transformation models for a single artistic style (\cite{Perceptual,18,32,33}) Synthesize stylized images using a single given artistic style image; (2) Training model that can convert multiple styles (\cite{3,7,35,20,37}) Introducing various network architectures while handling multiple styles; (3) Arbitrary style transformation model (\cite{12,21,microast,collaborative,meta,DynamicIN}) used different mechanisms such as feature modulation and matching to transfer any artistic style.

Looking back at all the above methods, , only Gayts (\cite{gatys}), DcDae (\cite{DcDae}), CTDP (\cite{CTDP}), and IDD (\cite{IDD}) can achieve high-quality texture transfer effects. Observing and analyzing the transfer results of CTDP in Fig.\ref{CTDP}, it is found that for the same style image, the texture parts in different generated results have extremely high similarity. Such high similarity texture features require encoding at deeper levels and a larger number of channels, so this operation is also the main source of the entire model's parameter count, computational complexity, and inference time. Therefore, although significant progress has been made in recent years, existing methods have overlooked the highly repetitive nature of texture features and still require repeated encoding for such redundant texture information.

\begin{figure*}[t]
	\centering
	\subfigure[CTDP]{\includegraphics[width=0.33\textwidth]{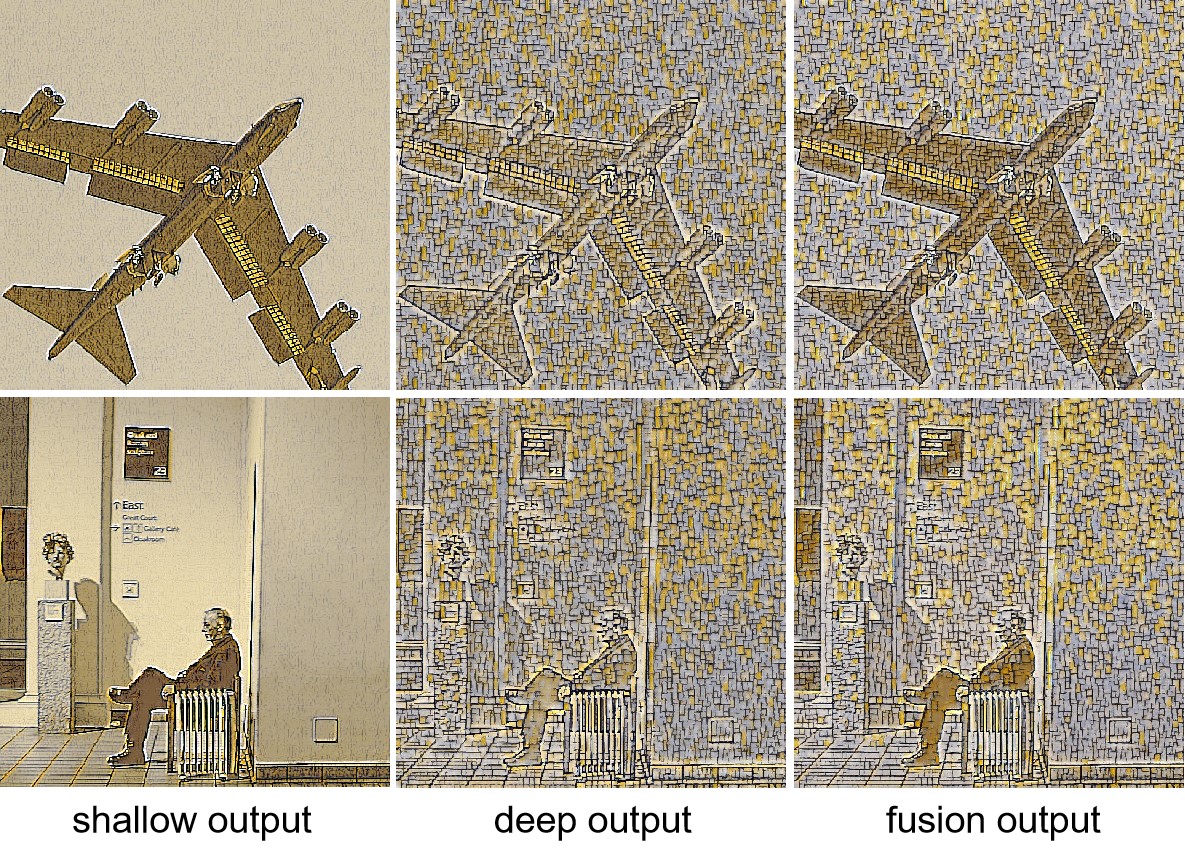}}
	\subfigure[CTDP (Pseudo texture feature preset method)]{\includegraphics[width=0.33\textwidth]{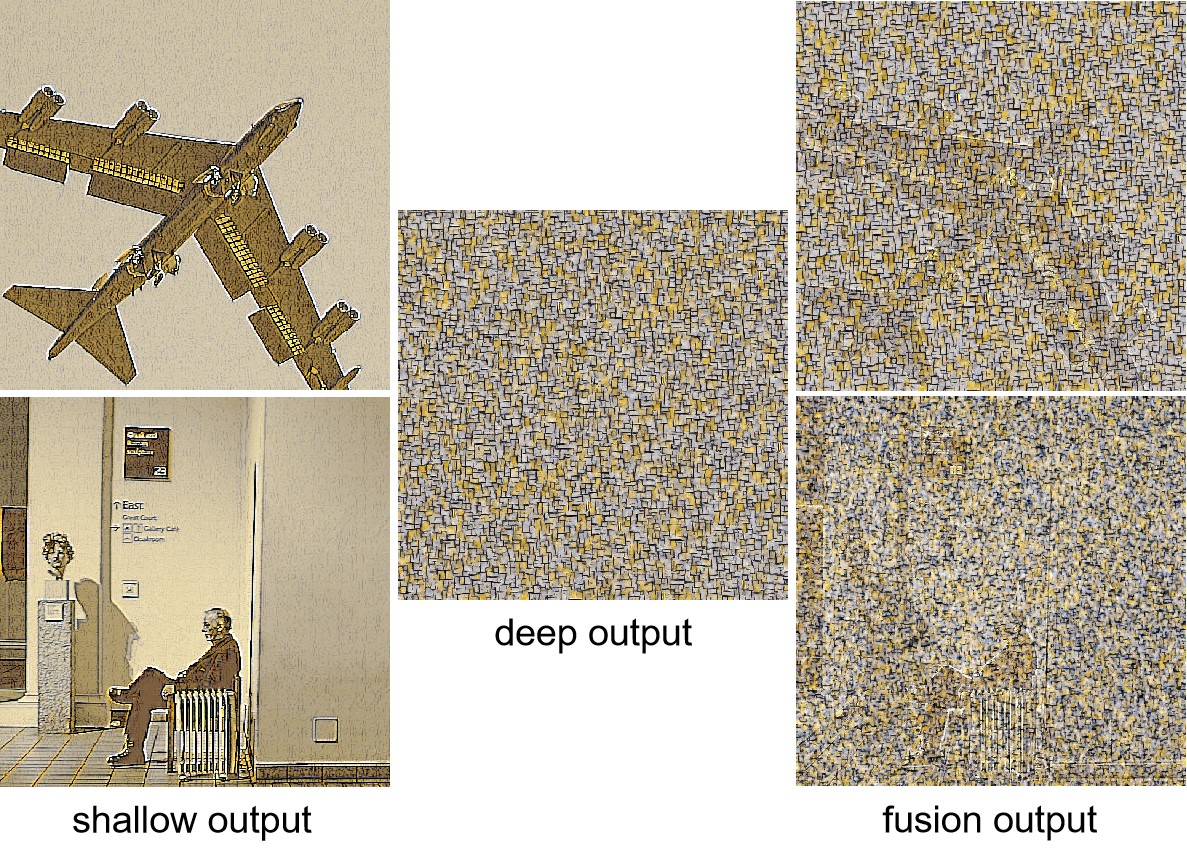}}
	\subfigure[TFP]{\includegraphics[width=0.33\textwidth]{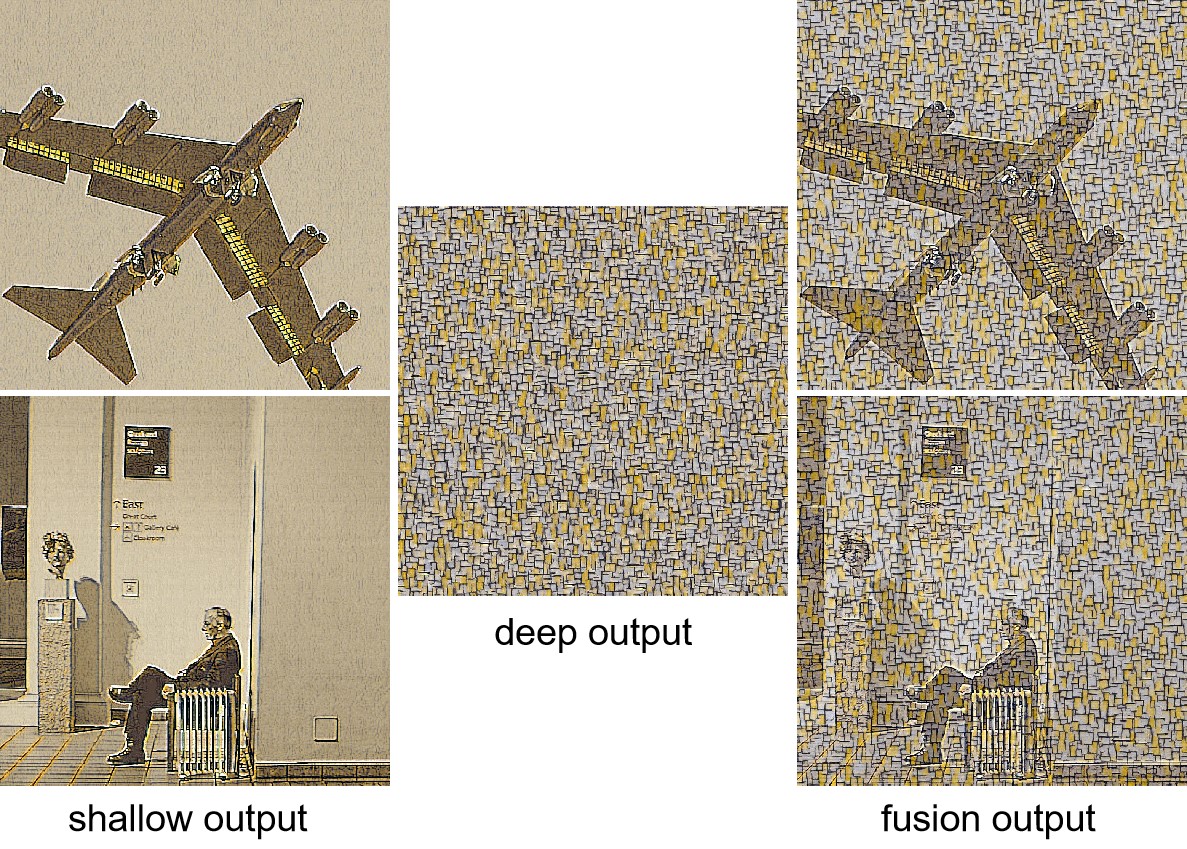}}
	\caption{\textbf{Ablation study of feature decoding consistency loss.} $\mathcal{L}_{mtv}$ ensures that shallow features in the shallow and fusion decoders' outputs yield similar results.}
	\label{ctdpftp}
\end{figure*}

In the face of the aforementioned challenges, we propose a lightweight texture transfer based on texture feature preset (\textbf{TFP}) model. This model can preset a well encoded universal deep texture feature map for a single style after training. In the inference stage, the preset texture feature map can be directly fused and decoded with shallow color transfer feature maps of any content, omitting the repeated encoding process for deep texture feature maps. On the basis of not changing the original framework of CTDP as much as possible, the model size of our texture feature preset scheme can be reduced by 3.2 times during the inference stage, and the inference speed can be accelerated by 1.8 times.

In addition to the improvements in model size and inference speed mentioned above, since the preset texture features are generated from noise, we can generate different texture feature maps such as Fig.\ref{random} in the inference stage by randomly sampling noise, thus generating different texture transfer results for the same content image. In addition, based on the input distribution differentiation experiment in IDD (\cite{IDD}), we have learned that the distribution differences within the input content image can lead to texture suppression differentiation performance issues. Similarly, we found a similar issue in the texture transfer task, where distribution differences within the same content image can lead to texture transfer differentiation, as shown in the red box area in Fig.\ref{wenlichayi}. However, in our method, deep texture feature maps are unconditionally generated pure texture images from noise images that completely follow the same noise distribution, which completely avoids the problem of texture transfer differentiated performance.

\begin{figure*}[t]
	\centering
	\includegraphics[width=\textwidth,height=\textheight,keepaspectratio]{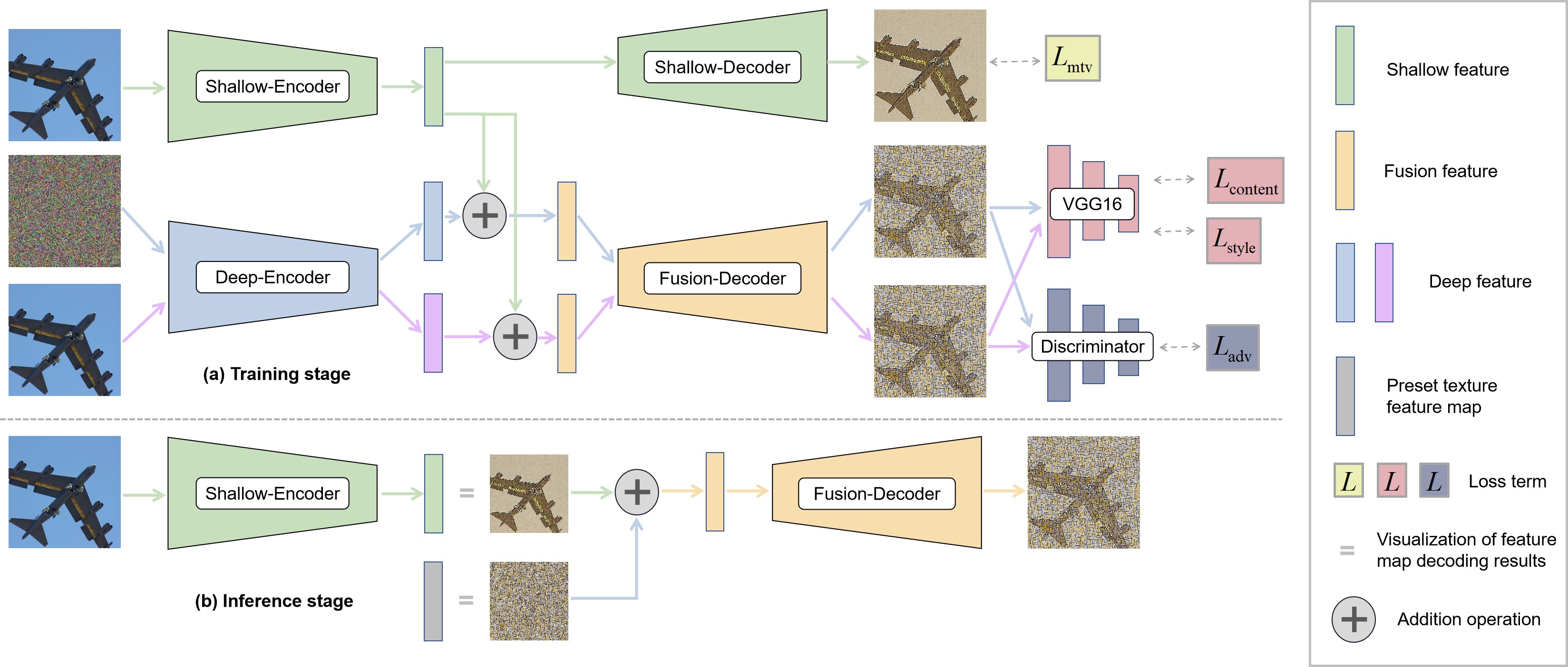}
	\caption{Architecture illustration of the proposed \textbf{CTDP}. See Section 3 for details.}
	\label{jiegou}
\end{figure*}

Compared to state-of-the-art models, our TFP not only produces visually superior results, but also has a volume that is 3.2-3538 times smaller and a speed that is 1.8-5.6 times faster. In summary, our contributions are as follows: 
\begin{itemize}
    \item We propose a lightweight texture transfer framework based on texture feature preset, which uses content independent noise input images to encode texture feature maps and fuse them with shallow feature maps of any content for decoding as the result of texture transfer.
    \item By presetting a deep texture feature map, we can directly skip the encoding process of the deep texture feature map during the inference stage, greatly reducing the model's parameter count, computational complexity, and inference speed.
    \item To prevent the semantic content from being completely masked by texture features, we designed a semantic noise texture fusion loss.
    \item To address the issue of local texture loss in texture feature maps caused by feature fusion decoding, we added a semantic conditional texture generation branch.
    \item The method of generating deep texture feature maps from noise can completely avoid the problem of texture transfer differentiation caused by the distribution differences within the content image.
    \item Due to the fact that texture features are generated by noise, random sampling of input noise can generate different texture feature maps and apply them to the texture transfer results.
    \item Numerous qualitative and quantitative experiments have shown that our method can quickly achieve high-quality texture transfer effects even with the fewest number of parameters.
\end{itemize} 

\section{Related work}
\subsection{Neural Style Transfer}
With the groundbreaking work of (\cite{gatys}), the era of neural style transfer (NST) has arrived. The visual appeal of style transfer has inspired subsequent researchers to improve in many aspects, including efficiency (\cite{Perceptual,32}); Quality (\cite{stroke,17,10,xie,DcDae}); Diversity (\cite{wang,chen}) and User Control (\cite{zhang,cham}); Despite significant progress, existing methods are difficult to process high-resolution images due to complex network structures and limited hardware resources.

\subsection{Lightweight Style Transfer}
To address the above challenges, (\cite{collaborative}) employed model compression techniques, known as collaborative distillation, to reduce the convolutional filters of VGG-19. While this method significantly reduced memory consumption, the pruned model was still not fast enough to run on 4K super-resolution images.(\cite{meta}) and (\cite{DynamicIN}) have designed lightweight networks, but still use pre-trained VGG models to extract style features, which can bring high computational costs and slow inference speed.

In order to achieve high-resolution style transfer, (\cite{chenchen}) divides the input image into small patches and use thumbnail instance normalization for patch-wise stylization to ensure style consistency between different patches. Although this method achieves 4K super-resolution style transfer, it essentially does not solve the problem of excessive forward inference time consumption.

\begin{figure*}[t]
	\centering
	\includegraphics[width=0.95\textwidth,height=\textheight,keepaspectratio]{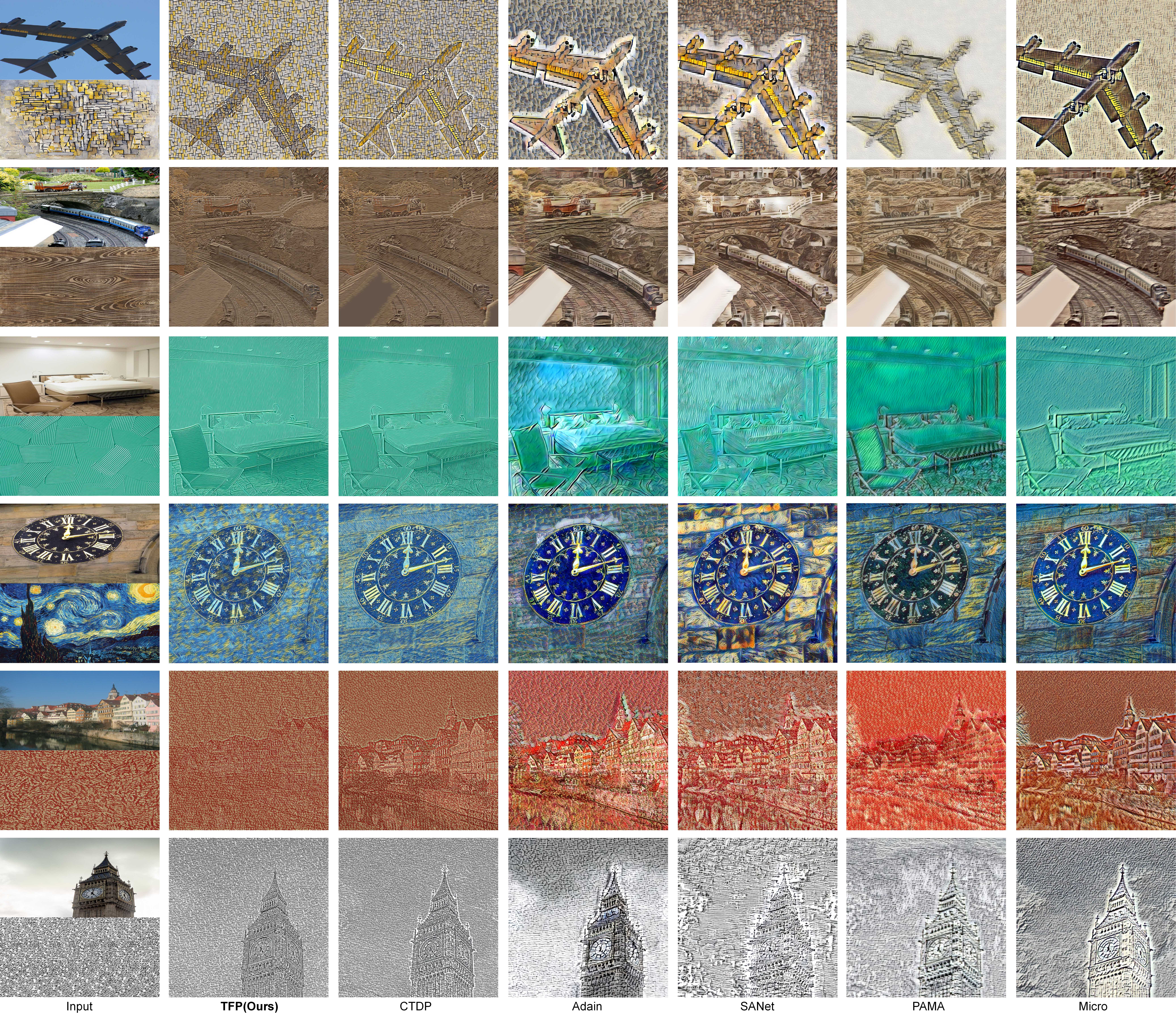}
	\caption{\textbf{Quantitative Comparison} with the state-of-the-art color and texture transfer methods using 1024 resolution input images. Due to the selection of many challenging style images with complex texture structures, it is best to zoom in to better observe artifact suppression and texture structure transfer.}
	\label{duibi}
\end{figure*}

Recently,(\cite{microast}) completely removed VGG and added a dual modulation strategy to inject color and texture structure information during the decoding phase. However, as shown in Fig.\ref{duibi}, experiments have shown that removing VGG style transfer significantly reduces the performance of arbitrary style transfer task, mainly manifested as color leakage, content structure distortion, and pseudo texture structure transfer. The transfer results for different texture structures are extremely similar, because the encoding and decoding of texture structures collapse into a unified compromise suboptimal texture structure.
\begin{figure}[t]
	\centering
	\includegraphics[width=0.5\textwidth,height=\textheight,keepaspectratio]{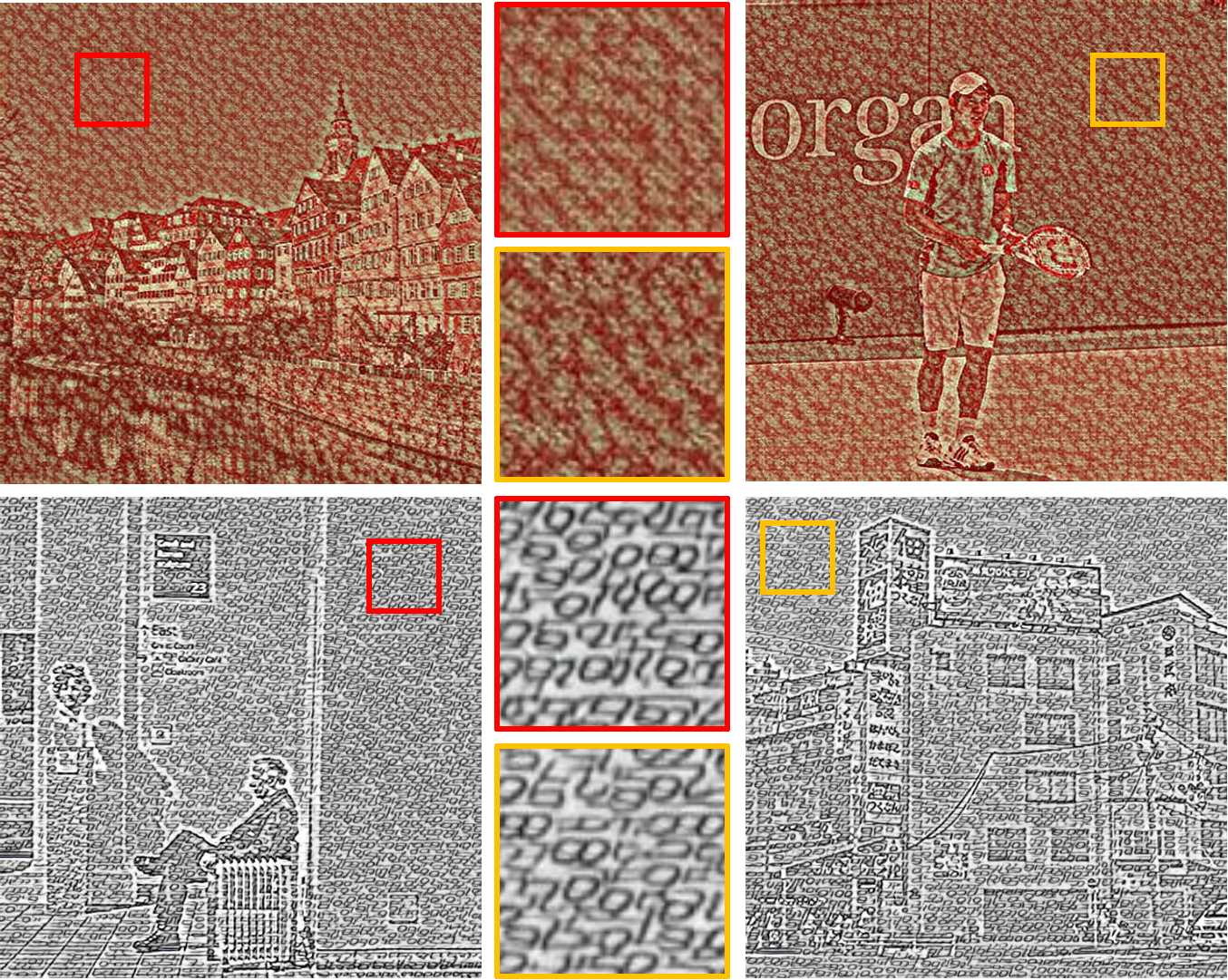}
	\caption{Architecture illustration of the proposed \textbf{CTDP}. See Section 3 for details.}
	\label{CTDP}
\end{figure}

\begin{figure*}[t]
	\centering
	\includegraphics[width=\textwidth,height=\textheight,keepaspectratio]{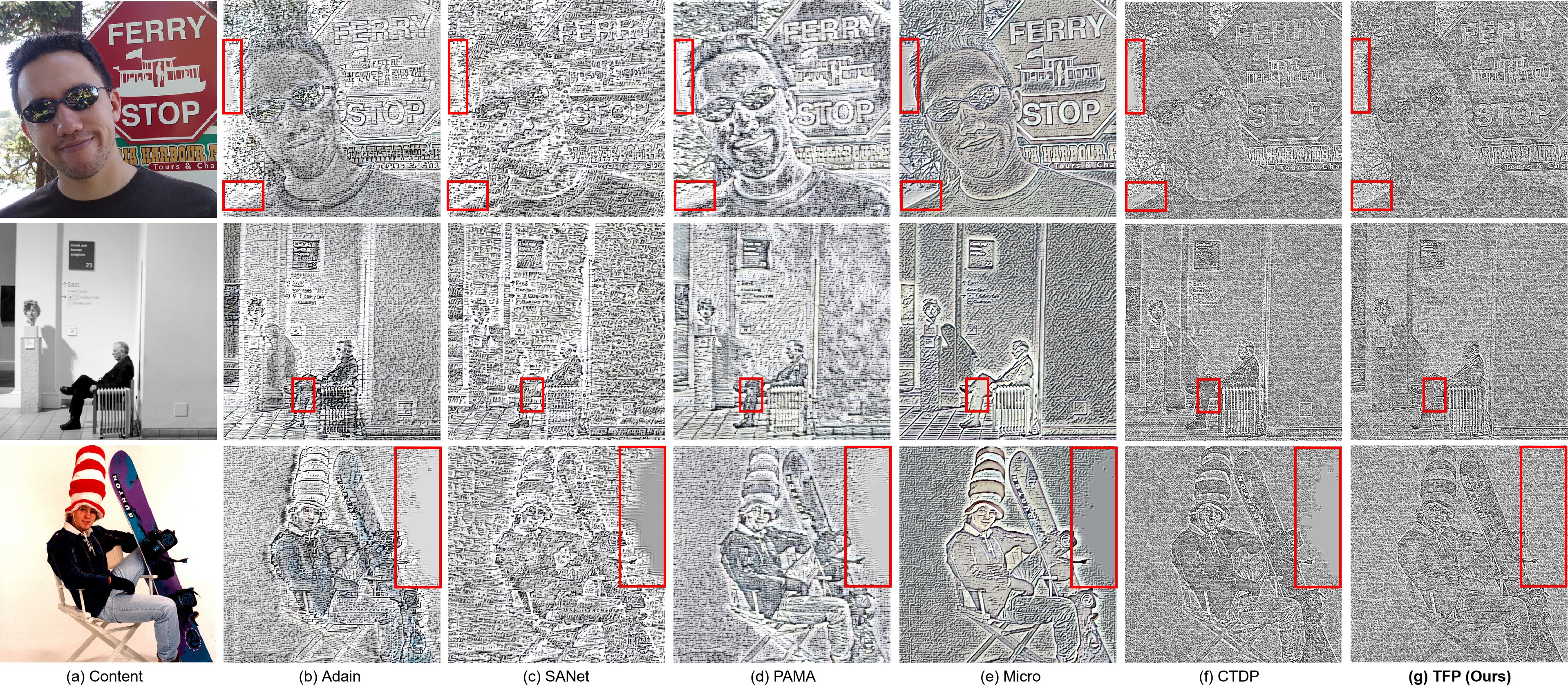}
	\caption{Architecture illustration of the proposed \textbf{CTDP}. See Section 3 for details.}
	\label{wenlichayi}
\end{figure*}

\begin{figure}[t]
	\centering
	\includegraphics[width=0.5\textwidth,height=\textheight,keepaspectratio]{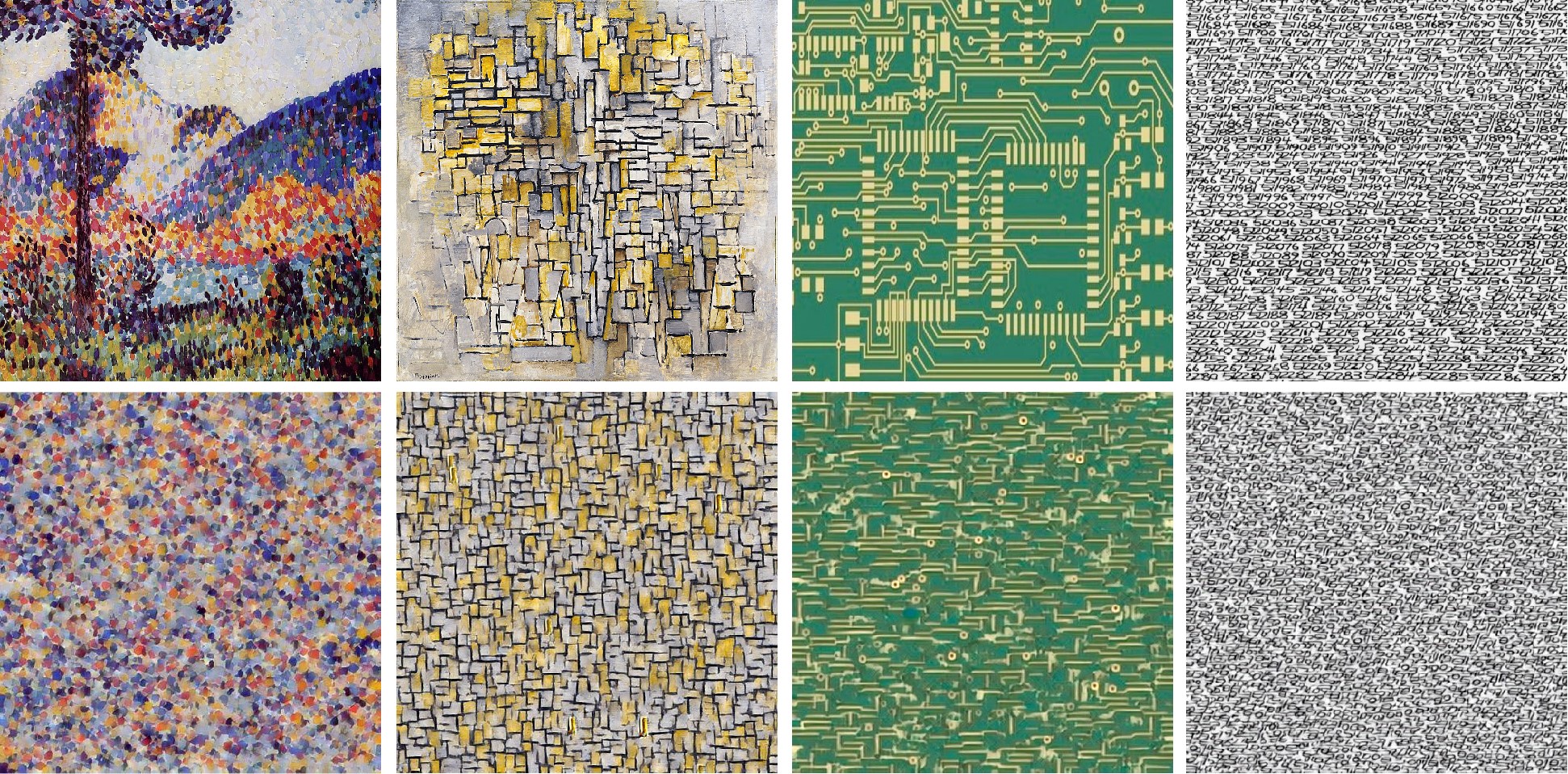}
	\caption{Architecture illustration of the proposed \textbf{CTDP}. See Section 3 for details.}
	\label{noise}
\end{figure}

\begin{figure*}[t]\centering
	\centering
	\includegraphics[width=\textwidth,height=\textheight,keepaspectratio]{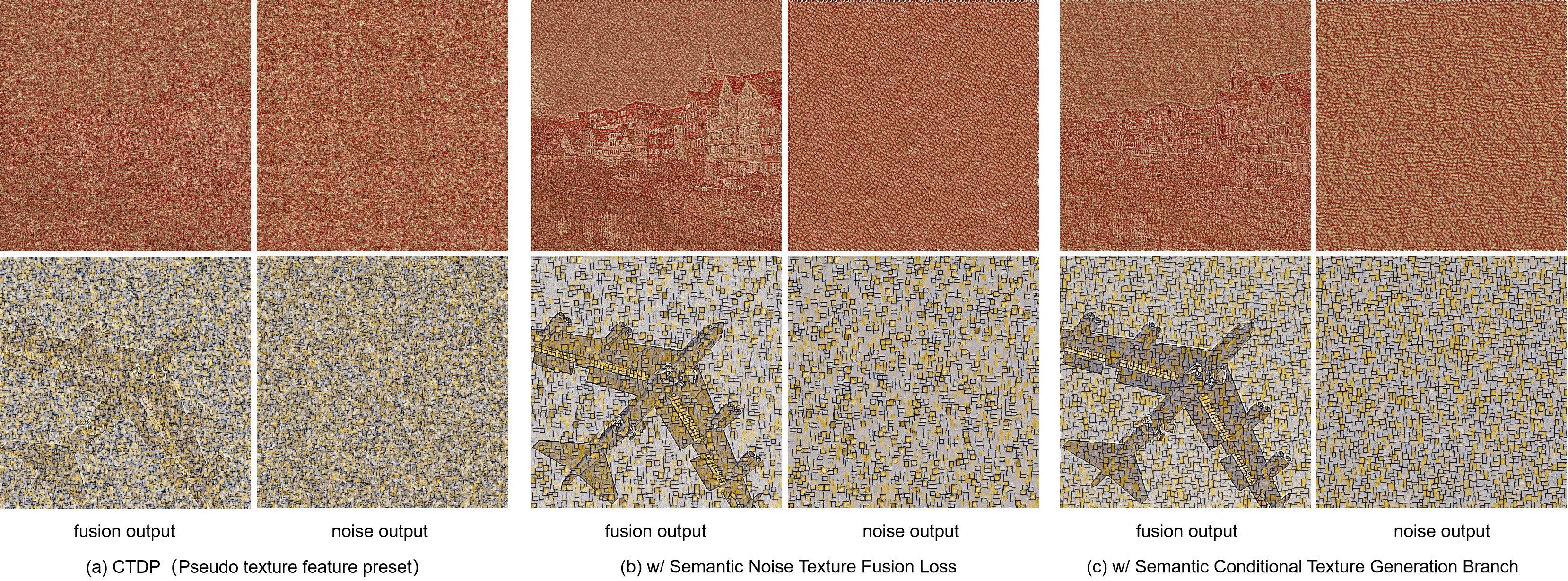}
	\caption{Visualization of feature maps for the first and third convolutions of three methods.}
	\label{method}
\end{figure*}

\section{Method}
Given an arbitrary content image, our goal is to achieve fast texture transfer through preset texture feature maps. The main challenges of this task lie in three aspects: (1) How to prevent semantic information from being completely masked by texture features during the fusion decoding process; (2) How to solve the problem of local texture missing in texture feature maps; (3) How to solve the problem of texture transfer differentiation;

\subsection{Overview of TFP}
As shown in Fig.\ref{jiegou}, our TFP framework consists of four main components: shallow encoder $Enc_s$ decoder $Dec_s$, deep encoder $Enc_d$, fusion decoder $Dec_f$, and style discriminator $D_s$ (only used during the training phase). Under this framework, $Enc_s$ are mainly responsible for encoding the semantics, details, and color transfer, while $Enc_d$ are mainly responsible for encoding deep texture feature maps from noisy inputs. Paired encoders and decoders have a symmetrical lightweight structure, consisting of two standard convolutional layers at the beginning and end, as well as several depthwise separable convolutional layers (DW, \cite{mobilenets}) in the middle. The complete forward inference pipeline of our framework is as follows:

(1) Extracting the shallow features $f^{c}_s$ of content image $C$ using a shallow encoder $Enc_s$, denoted as $f^{c}_s := Enc_s(C)$.

(2) Extracting the deep features $f^{n}_d$ of noise image $N$ using a deep encoder $Enc_d$, denoted as $f^{n}_d := Enc_d(N)$.

(3) Extracting the deep features $f^{c}_d$ of content image $C$ using a deep encoder $Enc_d$, denoted as $f^{c}_d := Enc_d(C)$.

(4) Obtain color transfer output $CS^{c}_ c$ by inputting shallow features $f^{c}_s$ into shallow decoder $Dec_s$, denoted as $CS^{c}_ c := Dec_s(f^{c}_s)$.
 
(5) Obtain noise texture transfer result $CS^{n}_ t$ by fusing and decoding the fusion features of $f^{c}_s$ and $f^{n}_d$, denoted as $CS^{n}_ t := Dec_f(\lambda_{s}Dae(f^{c}_s) + \lambda_{d}f^{n}_d)$.
 
(6) Obtain content texture transfer result $CS^{c}_ t$ by fusing and decoding the fusion features of $f^{c}_s$ and $f^{c}_d$, denoted as $CS^{c}_ t := Dec_f(\lambda_{s}Dae(f^{c}_s) + \lambda_{d}f^{c}_d)$.

Among them, $Dae$ represents the Detail Attention-enhanced (\cite{DcDae}) module, this framework primarily focuses on whether the input images are noise or content images. Therefore, we specifically denote the current input form using superscripts, where $c$ represents content input, and $n$ represents noise input. $\lambda_{s}$ and $\lambda_{d}$ represent the fusion strength of shallow and deep feature maps, respectively.

\textbf{Training Losses}. In order to achieve style transfer, similar to the previous method (\cite{gatys,xie,meta,microast,demystifying,SANet,adain,DcDae}), we use pre trained VGG-16 (\cite{vgg}) as our loss model to calculate content and style loss. We use perceptual loss (\cite{Perceptual}) as our branch content loss $\mathcal{L}_{bc}$, and all three of our branch content losses are calculated in the ${relu2\_1}$ layers of VGG-16. The branch style loss $\mathcal{L}_{bs}$ is defined as the matching Gram matrix (\cite{gatys}), and the three branches calculate the style loss at different levels (see details in Sec.\ref{sec:bs}). Introduce style discrimination loss $\mathcal{L}_{adv}$ similar to (\cite{DcDae}) to ensure the overall color and texture matching effect of stylized images. Please note that we only use VGG-16 during the training phase and do not require complex loss calculations or involve any large networks during the inference phase.
In summary, the overall goals of our TFP are:
\begin{equation}
	\label{zongloss}
	\mathcal{L}_{Full} = \lambda_{bc}\mathcal{L}_{bc} + \lambda_{bs}\mathcal{L}_{bs} + \lambda_{adv}\mathcal{L}_{adv} + \lambda_{mtv}\mathcal{L}_{mtv} + \lambda_{fdc}\mathcal{L}_{fdc} + \lambda_{stf}\mathcal{L}_{stf},
\end{equation}
where hyper-parameters $\lambda_{bc}$, $\lambda_{bs}$, $\lambda_{adv}$, $\lambda_{mtv}$ and $\lambda_{fdc}$ define the relative importance of each component in the total loss function.

\subsection{Background}
CTDP (\cite{CTDP}) pioneered the design of a dual pipeline framework for color and texture, which can simultaneously generate color and texture transfer results. As shown in Fig.\ref{ctdpftp}(a), CTDP actually produces three results simultaneously, namely shallow color transfer result, deep texture transfer result, and fusion decoding result. It generally takes the fusion decoding result as the final style transfer output.

\subsubsection{Texture Transfer Differentiation} We found that all previous schemes that required encoding of content images, including CTDP, and generating texture transfer results through semantic conditions all had texture transfer differentiation issues, as shown in the red box area in Fig.\ref{wenlichayi}. This phenomenon is similar to the texture suppression differentiation performance in IDD (\cite{IDD}), which is believed to be caused by the continuity of the input image. Discontinuous inputs will generate noise features in the feature map, and noise features will evolve into texture structures through convolution operations. Continuous inputs will not generate noise features in the feature map, and will not evolve into texture structure features. As shown in the first column of Fig.\ref{wenlichayi}, the extremely continuous areas in the content images directly do not generate any texture information, while the discontinuous parts outside the red box can effectively complete the texture transfer task. This phenomenon once again confirms the hypothesis of IDD.

\subsubsection{Texture Similarity} As shown in Fig.\ref{CTDP}, four output images of the CTDP model are displayed, with the top and bottom rows showing the stylized results of two different reference styles. By zooming in on the red and yellow box areas, we found that under the same reference style, the stylized output texture of different content images all have extremely high texture similarity. In the training and inference process of the model, generating such high similarity texture features requires more convolutional encoding, and more convolutional encoding means deeper network depth and larger number of channels, which makes the generation of texture features the main source of model parameters and computation. Is it necessary for us to repeatedly encode such redundant texture information?
\begin{figure}[t]
	\centering
	\includegraphics[width=0.5\textwidth,height=\textheight,keepaspectratio]{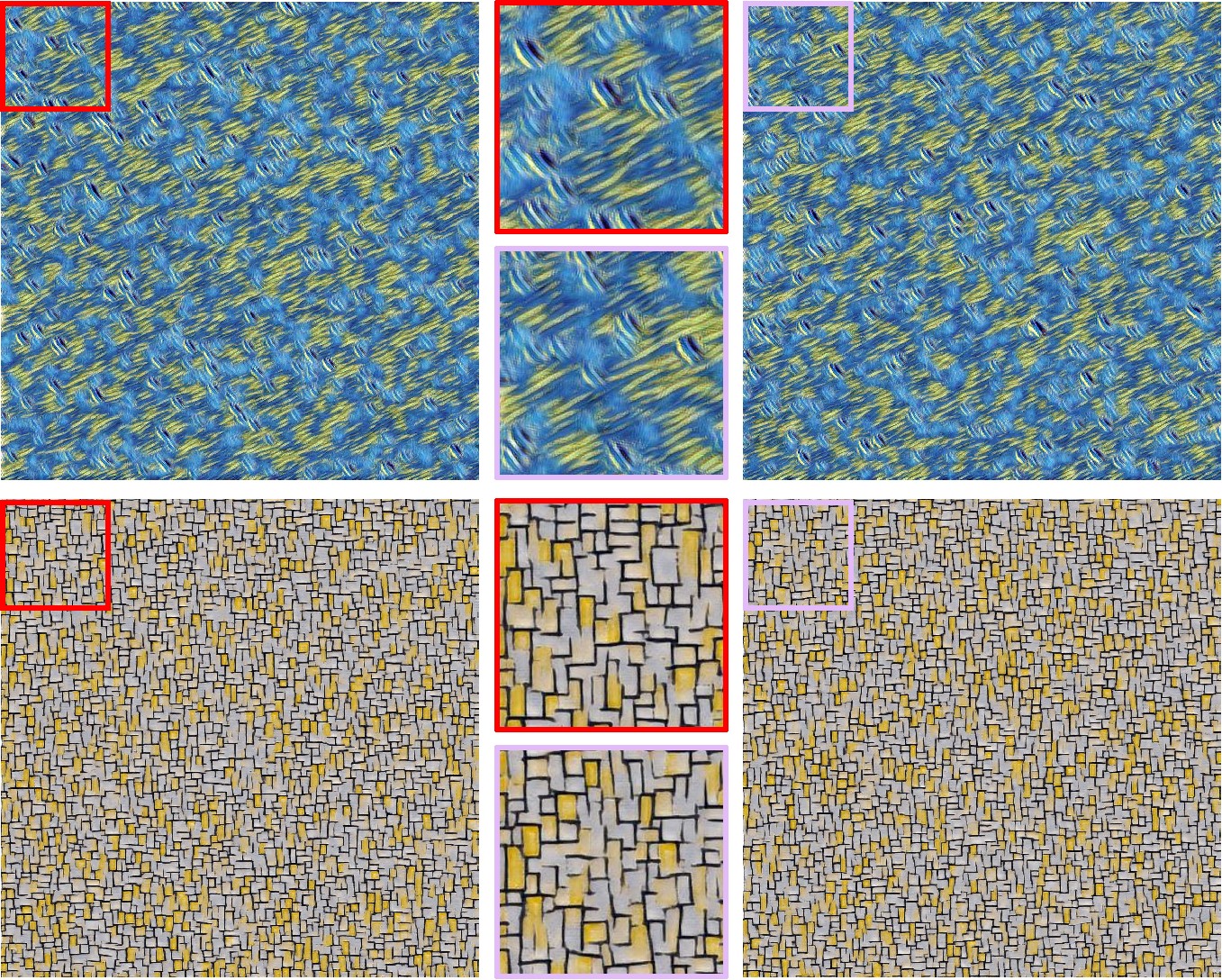}
	\caption{Architecture illustration of the proposed \textbf{CTDP}. See Section 3 for details.}
	\label{random}
\end{figure}

\subsubsection{Noise Input} We found in our experiment that replacing the input content image of CTDP with a pure noise image (standard normal distribution with mean 0 and variance 1) will generate a pure texture image as shown in the second row of Fig.\ref{noise}. The first row is its corresponding reference style image, that is, if the input is noise, the model will produce a pure texture image without any semantics. This phenomenon leads us to the following conjecture: 

(1) Due to the fact that only the Gram (\cite{gatys}) matrix is constrained for reference style images, the essence of such models is to perform color and texture reconstruction tasks; 

(2) When the input is a content image, it is essentially performing the task of conditional texture reconstruction, that is, content conditional generation;

(3) When the input is a content image, it essentially engages in the task of conditional texture reconstruction, namely, content conditional texture generation;

(4) If the result generated by unconditional generation is a pure texture image with no semantics or structure, can we achieve texture transfer by fitting the pure texture image onto the content image?

\subsubsection{Conclusion} Overall, our experiments on CTDP have yielded the following conclusions: 

(1) The distribution difference of input images can lead to texture transfer differentiation issues;

(2) Different content input images will produce texture transfer results with extremely high similarity in texture features; 

(3) The CTDP model essentially performs texture reconstruction tasks. When inputting noisy images, the model generates unconditionally generated pure texture images, and when inputting content images, the model generates conditionally generated texture transfer images with content semantics.

\subsection{Texture Feature Preset Framework}
We attempt to utilize the high repeatability of texture features and the model's ability to unconditionally generate pure texture images from noisy inputs to achieve texture feature preset (TFP) effects. TFP aims to provide preset texture feature maps for a single style, which can be fused and decoded with any shallow color transfer feature map to directly generate texture transfer results, thereby avoiding duplicate encoding of redundant texture information.

\subsubsection{Pseudo Texture Feature Preset}
Firstly, we attempt to directly execute the pseudo texture feature preset scheme on the pre-trained CTDP (\cite{CTDP}) framework. As shown in Fig.\ref{ctdpftp}(b), the three column outputs are the decoding results of shallow color transfer feature maps, noise feature map decoding, and fusion decoding of two feature maps. Observation shows that the pseudo TFP fusion decoding of the CTDP model results in a state where two images are directly superimposed, presenting an erroneous texture transfer effect where the content information is completely masked by texture features.

\subsubsection{Semantic Texture Fusion Loss}
We believe that the reason for the incorrect preset method of the pseudo texture features mentioned above is the lack of constraints on the noise encoded texture feature map. In fact, the pure texture feature map generated by noise in the previous experiment is only a side effect product of CTDP (\cite{CTDP}) framework training, and is not suitable for being directly used as the preset texture feature map.

To solve the fusion problem of shallow color transfer feature maps and deep noise texture feature maps, we designed a semantic texture fusion loss $\lambda{stf}$. Because we need to present the semantic and structural information of the reference content image and the texture features in the reference style image in the texture transfer results, $\lambda{stf}$ is actually designed based on the style and content perception loss of \cite{gatys} and \cite{Perceptual}. Unlike previous schemes that were calculated under the input of content images, our scheme calculates n * for randomly sampled noisy input images.

\begin{equation}
	\label{zongloss}
	\begin{aligned}
		CS^{n}_ t := Dec_f(\lambda_{s}Dae(Enc_s(c) ) + \lambda_{d}Enc_d(n))\\
	\end{aligned}
\end{equation}

\subsubsection{Semantic Conditional Texture Generation Branch}
As shown in the first row of Fig.\ref{method}(b), we did avoid the content semantics being masked by texture features through semantic texture fusion loss $\lambda{stf}$. However, when we observed the second row, we found that the texture map generated based on noise had a problem of local texture loss, which led to poor fusion decoding performance.

We believe it is the side effect of the direct constraint of loss $\lambda{stf}$ on noise that leads to the issue of local texture loss. Under the sole constraint of Loss $\lambda{stf}$, the model, in pursuit of the best semantic texture fusion effect, forces compromises in the encoding of noise-based texture feature maps, resulting in more significant local texture loss to better reduce Loss $\lambda{stf}$. Therefore, we introduced a semantic conditional texture generation branch, hoping that the semantic conditional encoding based on content images can guide the model in encoding deep texture feature maps beneficial for feature fusion.

As shown in Fig.\ref{method}(c), the top and bottom rows respectively show the predictive performance of two styles after training with the addition of semantic conditional texture generation branches. We observed that the color matching of the fusion decoding result in the first row is higher, the content color is almost not leaked, the artifacts in the noise output result are reduced, and the scale of the texture feature is increased. It presents an artistic effect where the content semantics are entirely composed of style texture patterns. The fusion decoding and texture feature decoding results in the second line have significantly solved the problem of local texture loss.

\subsection{Fast Texture Transfer}
The ultimate goal of our texture feature preset framework is to achieve faster inference speed in the inference stage by omitting repeated encoding of deep texture features. As shown in Fig.\ref{jiegou}(b), it is the execution process of TFP in the inference stage. The gray feature map is the preset deep texture feature map after training. We only need to perform shallow encoding on the content image $c \in R^{3 \times H \times W}$ to obtain the shallow color transfer feature map and fuse it with the preset texture feature map to decode and output the texture transfer result quickly, denoted as:

\begin{equation}
	\label{zongloss}
	CS^{n}_ t := Dec_f(\lambda_{s}Dae(Enc_s(c) ) + \lambda_{d}f^{n}_{tfp}).
\end{equation}
In this process, the shallow color transfer feature map is responsible for providing the structure and detail information of the content and completing the encoding of color transfer, while the preset texture feature map is responsible for providing complex and highly repetitive texture patterns with reference styles. As shown in Tab.\ref{compar}, TFP can achieve the fastest inference speed of 3.1ms for a single image at 256 resolution, which is 1.8 times faster than the previous fastest model.

\subsection{Random Texture Generation}

Since our deep texture feature maps are not encoded from content images, but from random noise maps, we can generate different texture feature maps during the inference stage by sampling different input noise and applying them to the texture transfer results. As shown in the Fig.\ref{random}, the upper and lower rows are pure texture images directly decoded from two styles of texture feature maps, and the two columns use different random sampling noise input images. The red and purple boxes are the enlarged results in the original image. We can observe that the texture patterns of the decoding results of texture feature maps of the same style are similar, but the combination and arrangement of features are not the same. This is the effect of different texture images with high similarity generated by random noise, and different texture feature maps can also be used to produce different texture transfer effects on a single content image through fusion decoding.

\subsection{Texture Transfer Differentiation}
As shown in the Fig.\ref{wenlichayi}, observing the red box in the figure, it can be found that all previous schemes did not achieve good texture transfer effects in the red box area, and there are serious texture transfer differentiation problems. This is because the background part of the content image has different degrees of continuity, and the red box in the image is different from other areas, which are extremely continuous and smooth. Previously, all solutions required texture encoding for content images, and there were often localized differences in the distribution of content images, especially in extremely continuous parts where the extremely continuous input parts could not generate texture representations, resulting in such texture transfer differentiation issues.

Our TFP scheme precisely avoids this content-based image based texture conditional encoding method, and instead relies entirely on an undifferentiated texture unconditional encoding method with the same noise distribution, generating a pure texture image with consistent global texture patterns. As shown in the last column of the Fig.\ref{wenlichayi}, the red box in our TFP scheme generates a highly consistent texture pattern with the other background parts.

\begin{table*}[t]
	\centering
	\captionsetup{position=bottom}
	\begin{tabular}{c|cccccc}
		\toprule
		Method&(a)Params/$10^6$&(b)Storage/MB&(c)GFLOPs&(d)Time/ms&(e)Prefer/\%(top1)&(f)Prefer/\%(top3)\\
		\midrule
		PAMA & 35.389 & 138.275 & 89.802 &  17.5& 0.13& 0.11\\
		SANet & 20.911 & 81.703 & 66.924 & 9.2&  0.06& 0.05\\
		Adain & 7.01 & 27.397 & 47.459 & 6.2&  0.06& 0.05\\
		Micro & 0.472 & 1.866 & 2.765 & 5.5&  0.06& 0.05\\
		CTDP & 0.032 &  0.15& 0.935 & 5.5 & 0.17& 0.15\\
		\midrule
		\textbf{TFP(Ours)} & \textbf{ 0.01} & \textbf{0.05}& \textbf{0.545}& \textbf{3.1}& \textbf{0.36}& \textbf{0.48}\\
		\textbf{TFP-L(Ours)} & \textbf{ 0.007} & \textbf{0.039}& \textbf{0.398}& \textbf{2.3}& \textbf{-}& \textbf{-}\\
		\bottomrule
	\end{tabular}
	\caption{\textbf{Quantitative Comparison} with State-of-the-Art Methods. Storage space is measured within the PyTorch model. GFLOPs and time are measured when both content and style are 4K images, and tested on the NVIDIA 3090 (24GB) GPU.The best results are highlighted in bold. OOM: Out of memory.} 
	\label{compar}
\end{table*} 

\section{Experiments}
\subsection{Implementation Details}
We used MS-COO (\cite{coco}) as the content image and extracted style images from Wikiart (\cite{wiki}) to train our TFP model. In equation.\ref{zongloss}, the values of $\lambda_{bc}$,  $\lambda_{bs}$, $\lambda_{adv}$, $\lambda_{mtv}$, $\lambda_{fdc}$ and $\lambda_{stf}$ are set to 1e0, 1e5, 1e0, 2e-5, 1e0 and 1e0, respectively. We used the Adam (\cite{adam}) optimizer with a learning rate of 0.001. During the training process, first adjust the size of the content image to 512, and then randomly crop it to 256 $\times$ 256 pixels for enhancement. Style images are processed using similar methods, but all images in a batch are randomly cropped from the same reference style image. Unlike the previous plan, we also need to randomly sample a batch of random noise with a size of 256 $\times$ 256 as input. We conducted all experiments on the RTX 3090 GPU.
\subsection{Comparisons with Prior Arts}
Due to our model's ability to quickly generate color and texture transfer results simultaneously, we compared our CTDP with state-of-the-art color transfer models and texture transfer models (arbitrary style transfer). In the comparison scheme, we directly ran the code with the default settings published by the author.

\begin{figure}[t]\centering
	
	\subfigure[Full Model]{\includegraphics[width=0.156\textwidth]{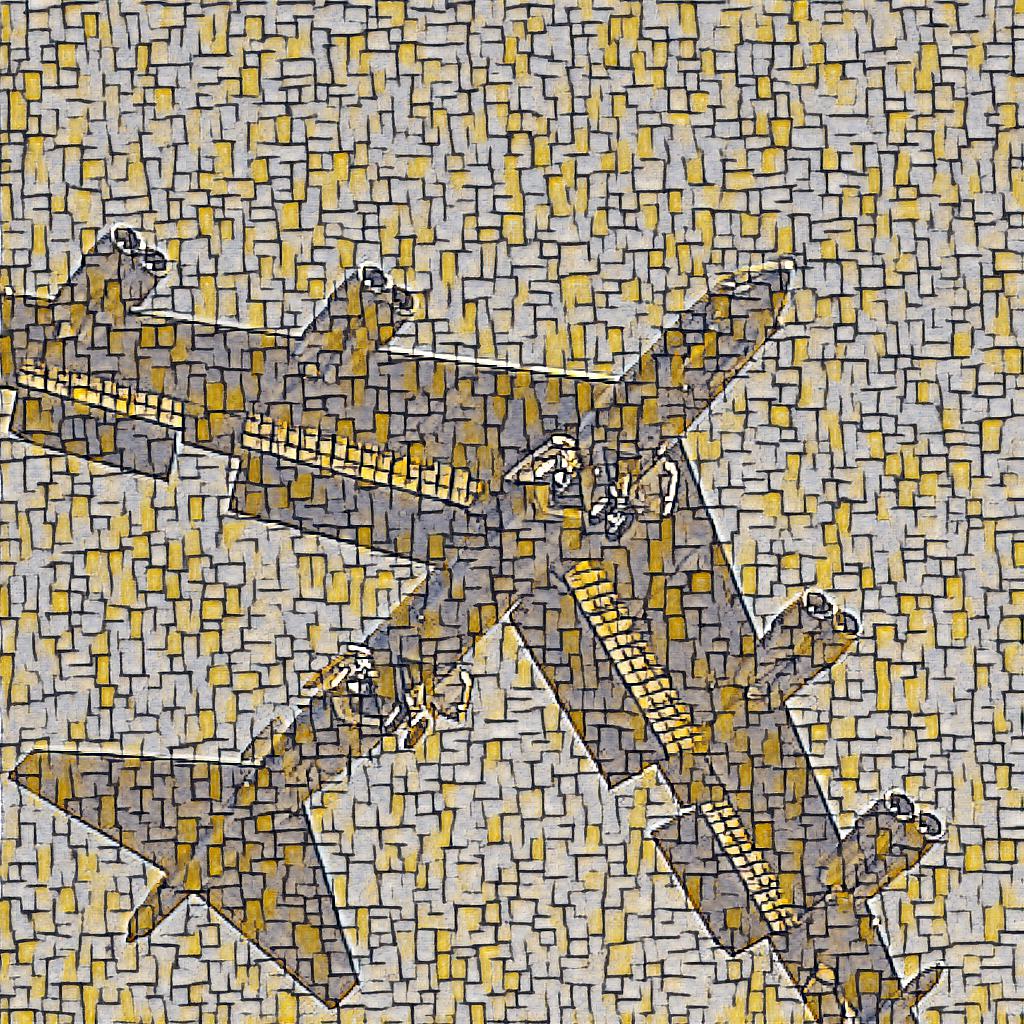}}
	\subfigure[w/o $\mathcal{L}_{bs}$]{\includegraphics[width=0.156\textwidth]{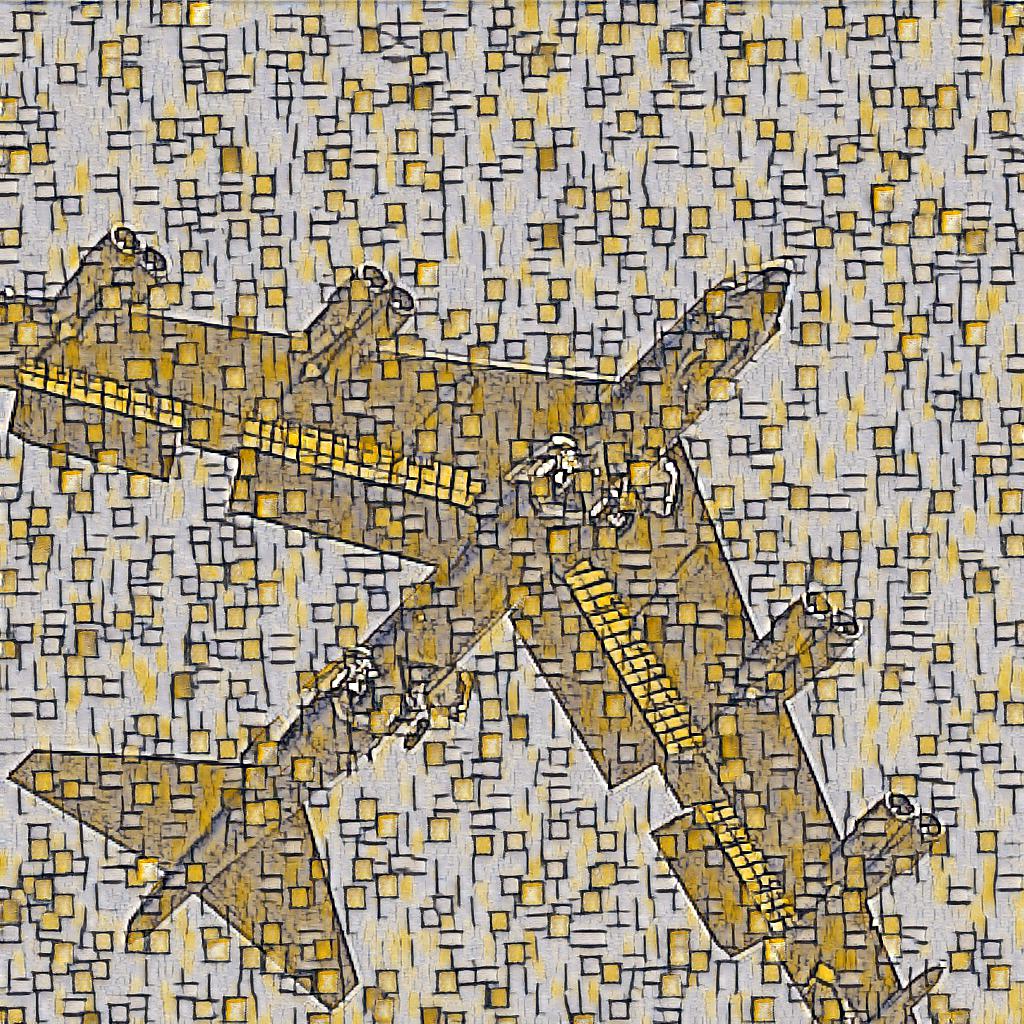}}
	\subfigure[w/o $\mathcal{L}_{mtv}$]{\includegraphics[width=0.156\textwidth]{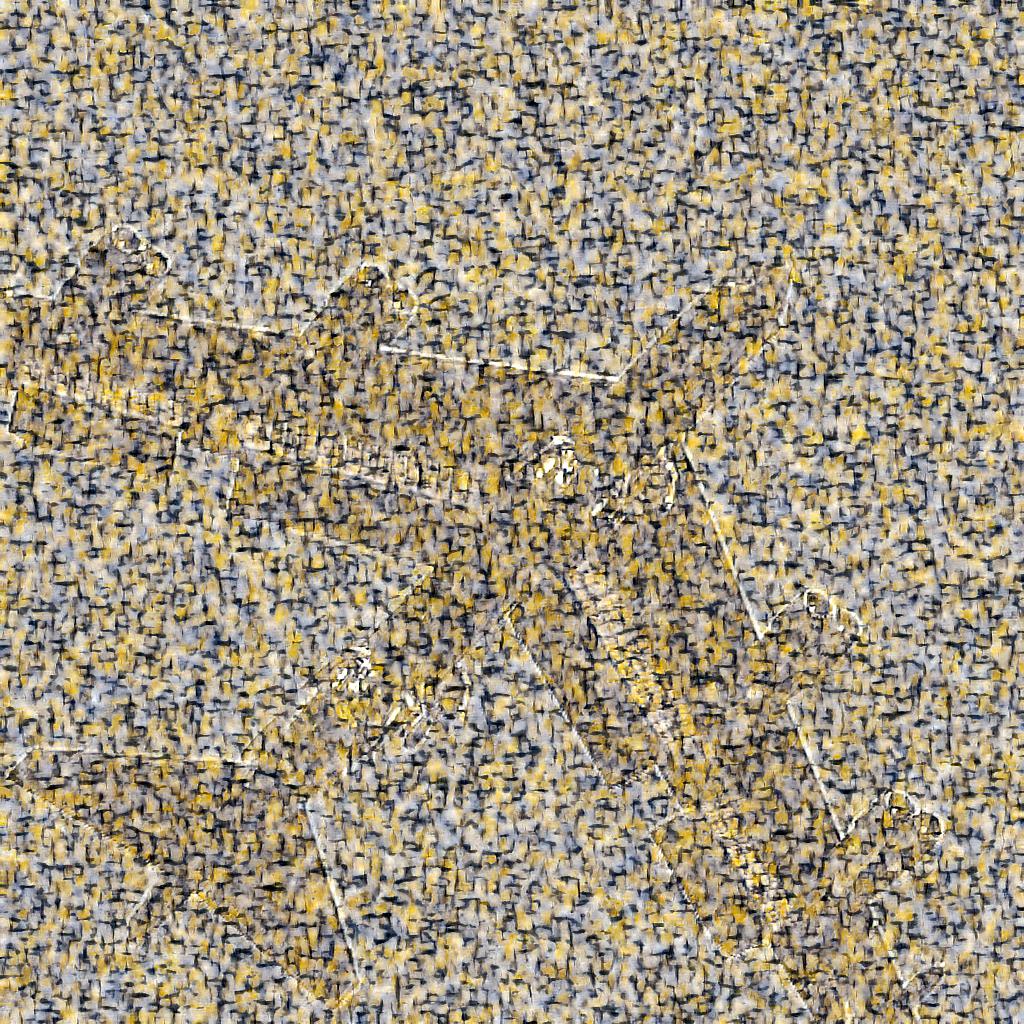}}
	\caption{\textbf{Ablation study} of branch style loss and masked total variation loss  to evaluate their effectiveness in suppressing texture and artifacts in color transfer tasks.}
	\label{A}
\end{figure}

\subsubsection{Qualitative Comparison}
The qualitative comparison results of different texture transfer methods are shown in Fig.\ref{duibi}.

Firstly, compare with our most relevant work, CTDP (\cite{CTDP}). In the CTDP results of the first, fifth, and sixth rows of the Fig.\ref{duibi}, it can be observed that there is a halo problem at the semantic edges of the aircraft, house, and clock tower. TFP has to some extent solved this problem, and there is no obvious or conflicting edge halo. In the second row of CTDP results in the Fig.\ref{duibi}, there is a significant texture transfer differentiation problem in the lower left corner. The smoother content input results in CTDP not encoding texture information on it, while TFP avoids this problem. TFP achieves an effect comparable to CTDP in terms of matching texture structure information and color information.

Adain (\cite{adain}), SANet (\cite{SANet}), PAMA (\cite{pama}), and Micro (\cite{microast}) all have significant issues with texture transfer quality. In the second row of the Fig.\ref{duibi}, all schemes in the bottom left corner have obvious texture transfer differentiation issues. Although PAMA and Micro schemes completely lose their texture here, they at least have the effect of color transfer. Adain and SANet even failed to encode the color, completely leaking the background color of the content. In the results of all the schemes in the first, third, fifth, and sixth rows of the Fig.\ref{duibi}, it can be observed that there are very obvious halo problems at the edges of the main objects of the airplane, bed and chair, house, and clock tower. The texture matching degree in all results of Adain and Micro schemes is very low. In Micro, all results have similar texture patterns and do not correspond to the reference style. There are different texture patterns in Adain, but the matching degree with the reference texture pattern is not high. PAMA and SANet only match the texture patterns correctly in the third and fourth rows, but their color and overall migration effects are slightly inferior. The texture and color matching degree in other reference images are not high, and the migration effect is poor.

In contrast, our TFP achieves state-of-the-art texture transfer effects. The global color and texture structure of all our results have a high degree of matching, and it is the only solution that can avoid the problem of subject object edge halo and texture transfer differentiation.

\subsubsection{Quantitative Comparison} Tab.\ref{compar} shows the quantitative comparison between our model and state-of-the-art methods in the inference stage. Due to the lack of widely accepted quantitative evaluation metrics for style transfer tasks in the industry, we only compared model size and forward inference speed in this study. As shown in columns a and d of Tab.\ref{compar}, our TFP is 3.2-3538 times smaller and 1.8-5.6 times faster than existing models.

\subsubsection{User Study}The evaluation of stylized results is highly subjective. Therefore, we conducted user studies on these five methods. We randomly presented 30 randomly shuffled stylized images to each participant, and each of the six methods (CTDP, AdaIN, PAMA, MicroAST, SANet et al, and ours) generated five stylized images, which were then shuffled randomly. Participants had unlimited time to select their favorite image (top 1) and top three images (top 3). We collected a total of 108 valid votes (three per person) from 36 participants, and show the percentage of preferred results for each method in the last two columns of Table for top 1 and top 3, respectively. Finally, as shown in Tab.\ref{compar}, the results indicate that our stylized images are more attractive than those of competitors.

\subsection{Ablation Study}
The result without content texture fusion loss is shown in Fig.\ref{A}(b), where the semantic information of the content image is completely masked by texture features, and the shallow color transfer feature map and deep texture feature map are not well fused and decoded.

The result without semantic conditional texture encoding branch is shown in Fig.\ref{A}(c), and there is a problem of local texture loss in the texture part of the result.

\section{Conclusion}
In this article, we propose a dual pipeline lightweight framework called CTDP. For the first time, our dual channels can simultaneously generate color and texture transfer results corresponding to style images, and the weighted fusion of dual branch features achieves the effect of adding texture features with controllable intensity from color transfer results for the first time. In addition, mtv loss was designed to suppress texture information in the model matching Gram matrix, and it was found that smoothing the input in our framework can almost completely eliminate texture features. A large number of experiments have proven the effectiveness of this method. Compared to the current level of technology, our CTDP is the first model that can simultaneously achieve color and texture transfer. It not only produces visually superior results in both migration tasks, but also has a color migration branch model size as low as 20k.

\bibliographystyle{elsarticle-harv}
\bibliography{ref}

\begin{thebibliography}{39}
\expandafter\ifx\csname natexlab\endcsname\relax\def\natexlab#1{#1}\fi
\providecommand{\url}[1]{\texttt{#1}}
\providecommand{\href}[2]{#2}
\providecommand{\path}[1]{#1}
\providecommand{\DOIprefix}{doi:}
\providecommand{\ArXivprefix}{arXiv:}
\providecommand{\URLprefix}{URL: }
\providecommand{\Pubmedprefix}{pmid:}
\providecommand{\doi}[1]{\href{http://dx.doi.org/#1}{\path{#1}}}
\providecommand{\Pubmed}[1]{\href{pmid:#1}{\path{#1}}}
\providecommand{\bibinfo}[2]{#2}
\ifx\xfnm\relax \def\xfnm[#1]{\unskip,\space#1}\fi
%Type = Article
\bibitem[{Champandard(2016)}]{cham}
\bibinfo{author}{Champandard, A.J.}, \bibinfo{year}{2016}.
\newblock \bibinfo{title}{Semantic style transfer and turning two-bit doodles
  into fine artworks}.
\newblock \bibinfo{journal}{arXiv preprint arXiv:1603.01768} .
%Type = Inproceedings
\bibitem[{Chen et~al.(2017)Chen, Yuan, Liao, Yu and Hua}]{3}
\bibinfo{author}{Chen, D.}, \bibinfo{author}{Yuan, L.}, \bibinfo{author}{Liao,
  J.}, \bibinfo{author}{Yu, N.}, \bibinfo{author}{Hua, G.},
  \bibinfo{year}{2017}.
\newblock \bibinfo{title}{Stylebank: An explicit representation for neural
  image style transfer}, in: \bibinfo{booktitle}{Proceedings of the IEEE
  conference on computer vision and pattern recognition}, pp.
  \bibinfo{pages}{1897--1906}.
%Type = Inproceedings
\bibitem[{Chen et~al.(2021)Chen, Zhao, Zhang, Wang, Zuo, Li, Xing and
  Lu}]{chen}
\bibinfo{author}{Chen, H.}, \bibinfo{author}{Zhao, L.}, \bibinfo{author}{Zhang,
  H.}, \bibinfo{author}{Wang, Z.}, \bibinfo{author}{Zuo, Z.},
  \bibinfo{author}{Li, A.}, \bibinfo{author}{Xing, W.}, \bibinfo{author}{Lu,
  D.}, \bibinfo{year}{2021}.
\newblock \bibinfo{title}{Diverse image style transfer via invertible
  cross-space mapping}, in: \bibinfo{booktitle}{2021 IEEE/CVF International
  Conference on Computer Vision (ICCV)}, \bibinfo{organization}{IEEE Computer
  Society}. pp. \bibinfo{pages}{14860--14869}.
%Type = Inproceedings
\bibitem[{Chen et~al.(2022)Chen, Wang, Xie, Lu and Luo}]{chenchen}
\bibinfo{author}{Chen, Z.}, \bibinfo{author}{Wang, W.}, \bibinfo{author}{Xie,
  E.}, \bibinfo{author}{Lu, T.}, \bibinfo{author}{Luo, P.},
  \bibinfo{year}{2022}.
\newblock \bibinfo{title}{Towards ultra-resolution neural style transfer via
  thumbnail instance normalization}, in: \bibinfo{booktitle}{Proceedings of the
  AAAI Conference on Artificial Intelligence}, pp. \bibinfo{pages}{393--400}.
%Type = Article
\bibitem[{Dumoulin et~al.(2016)Dumoulin, Shlens and Kudlur}]{7}
\bibinfo{author}{Dumoulin, V.}, \bibinfo{author}{Shlens, J.},
  \bibinfo{author}{Kudlur, M.}, \bibinfo{year}{2016}.
\newblock \bibinfo{title}{A learned representation for artistic style}.
\newblock \bibinfo{journal}{arXiv preprint arXiv:1610.07629} .
%Type = Inproceedings
\bibitem[{Gatys et~al.(2016)Gatys, Ecker and Bethge}]{gatys}
\bibinfo{author}{Gatys, L.A.}, \bibinfo{author}{Ecker, A.S.},
  \bibinfo{author}{Bethge, M.}, \bibinfo{year}{2016}.
\newblock \bibinfo{title}{Image style transfer using convolutional neural
  networks}, in: \bibinfo{booktitle}{Proceedings of the IEEE conference on
  computer vision and pattern recognition}, pp. \bibinfo{pages}{2414--2423}.
%Type = Inproceedings
\bibitem[{Gu et~al.(2018)Gu, Chen, Liao and Yuan}]{10}
\bibinfo{author}{Gu, S.}, \bibinfo{author}{Chen, C.}, \bibinfo{author}{Liao,
  J.}, \bibinfo{author}{Yuan, L.}, \bibinfo{year}{2018}.
\newblock \bibinfo{title}{Arbitrary style transfer with deep feature
  reshuffle}, in: \bibinfo{booktitle}{Proceedings of the IEEE Conference on
  Computer Vision and Pattern Recognition}, pp. \bibinfo{pages}{8222--8231}.
%Type = Article
\bibitem[{Howard et~al.(2017)Howard, Zhu, Chen, Kalenichenko, Wang, Weyand,
  Andreetto and Adam}]{mobilenets}
\bibinfo{author}{Howard, A.G.}, \bibinfo{author}{Zhu, M.},
  \bibinfo{author}{Chen, B.}, \bibinfo{author}{Kalenichenko, D.},
  \bibinfo{author}{Wang, W.}, \bibinfo{author}{Weyand, T.},
  \bibinfo{author}{Andreetto, M.}, \bibinfo{author}{Adam, H.},
  \bibinfo{year}{2017}.
\newblock \bibinfo{title}{Mobilenets: Efficient convolutional neural networks
  for mobile vision applications}.
\newblock \bibinfo{journal}{arXiv preprint arXiv:1704.04861} .
%Type = Inproceedings
\bibitem[{Huang and Belongie(2017)}]{adain}
\bibinfo{author}{Huang, X.}, \bibinfo{author}{Belongie, S.},
  \bibinfo{year}{2017}.
\newblock \bibinfo{title}{Arbitrary style transfer in real-time with adaptive
  instance normalization}, in: \bibinfo{booktitle}{Proceedings of the IEEE
  international conference on computer vision}, pp.
  \bibinfo{pages}{1501--1510}.
%Type = Inproceedings
\bibitem[{Jing et~al.(2020)Jing, Liu, Ding, Wang, Ding, Song and
  Wen}]{DynamicIN}
\bibinfo{author}{Jing, Y.}, \bibinfo{author}{Liu, X.}, \bibinfo{author}{Ding,
  Y.}, \bibinfo{author}{Wang, X.}, \bibinfo{author}{Ding, E.},
  \bibinfo{author}{Song, M.}, \bibinfo{author}{Wen, S.}, \bibinfo{year}{2020}.
\newblock \bibinfo{title}{Dynamic instance normalization for arbitrary style
  transfer}, in: \bibinfo{booktitle}{Proceedings of the AAAI Conference on
  Artificial Intelligence}, pp. \bibinfo{pages}{4369--4376}.
%Type = Inproceedings
\bibitem[{Jing et~al.(2018)Jing, Liu, Yang, Feng, Yu, Tao and Song}]{stroke}
\bibinfo{author}{Jing, Y.}, \bibinfo{author}{Liu, Y.}, \bibinfo{author}{Yang,
  Y.}, \bibinfo{author}{Feng, Z.}, \bibinfo{author}{Yu, Y.},
  \bibinfo{author}{Tao, D.}, \bibinfo{author}{Song, M.}, \bibinfo{year}{2018}.
\newblock \bibinfo{title}{Stroke controllable fast style transfer with adaptive
  receptive fields}, in: \bibinfo{booktitle}{Proceedings of the European
  Conference on Computer Vision (ECCV)}, pp. \bibinfo{pages}{238--254}.
%Type = Inproceedings
\bibitem[{Johnson et~al.(2016)Johnson, Alahi and Fei-Fei}]{Perceptual}
\bibinfo{author}{Johnson, J.}, \bibinfo{author}{Alahi, A.},
  \bibinfo{author}{Fei-Fei, L.}, \bibinfo{year}{2016}.
\newblock \bibinfo{title}{Perceptual losses for real-time style transfer and
  super-resolution}, in: \bibinfo{booktitle}{Computer Vision--ECCV 2016: 14th
  European Conference, Amsterdam, The Netherlands, October 11-14, 2016,
  Proceedings, Part II 14}, \bibinfo{organization}{Springer}. pp.
  \bibinfo{pages}{694--711}.
%Type = Article
\bibitem[{Kingma and Ba(2014)}]{adam}
\bibinfo{author}{Kingma, D.P.}, \bibinfo{author}{Ba, J.}, \bibinfo{year}{2014}.
\newblock \bibinfo{title}{Adam: A method for stochastic optimization}.
\newblock \bibinfo{journal}{arXiv preprint arXiv:1412.6980} .
%Type = Inproceedings
\bibitem[{Kolkin et~al.(2019)Kolkin, Salavon and Shakhnarovich}]{15}
\bibinfo{author}{Kolkin, N.}, \bibinfo{author}{Salavon, J.},
  \bibinfo{author}{Shakhnarovich, G.}, \bibinfo{year}{2019}.
\newblock \bibinfo{title}{Style transfer by relaxed optimal transport and
  self-similarity}, in: \bibinfo{booktitle}{Proceedings of the IEEE/CVF
  Conference on Computer Vision and Pattern Recognition}, pp.
  \bibinfo{pages}{10051--10060}.
%Type = Inproceedings
\bibitem[{Li and Wand(2016a)}]{17}
\bibinfo{author}{Li, C.}, \bibinfo{author}{Wand, M.}, \bibinfo{year}{2016}a.
\newblock \bibinfo{title}{Combining markov random fields and convolutional
  neural networks for image synthesis}, in: \bibinfo{booktitle}{Proceedings of
  the IEEE conference on computer vision and pattern recognition}, pp.
  \bibinfo{pages}{2479--2486}.
%Type = Inproceedings
\bibitem[{Li and Wand(2016b)}]{18}
\bibinfo{author}{Li, C.}, \bibinfo{author}{Wand, M.}, \bibinfo{year}{2016}b.
\newblock \bibinfo{title}{Precomputed real-time texture synthesis with
  markovian generative adversarial networks}, in: \bibinfo{booktitle}{Computer
  Vision--ECCV 2016: 14th European Conference, Amsterdam, The Netherlands,
  October 11-14, 2016, Proceedings, Part III 14},
  \bibinfo{organization}{Springer}. pp. \bibinfo{pages}{702--716}.
%Type = Inproceedings
\bibitem[{Li et~al.(2017a)Li, Fang, Yang, Wang, Lu and Yang}]{20}
\bibinfo{author}{Li, Y.}, \bibinfo{author}{Fang, C.}, \bibinfo{author}{Yang,
  J.}, \bibinfo{author}{Wang, Z.}, \bibinfo{author}{Lu, X.},
  \bibinfo{author}{Yang, M.H.}, \bibinfo{year}{2017}a.
\newblock \bibinfo{title}{Diversified texture synthesis with feed-forward
  networks}, in: \bibinfo{booktitle}{Proceedings of the IEEE conference on
  computer vision and pattern recognition}, pp. \bibinfo{pages}{3920--3928}.
%Type = Article
\bibitem[{Li et~al.(2017b)Li, Fang, Yang, Wang, Lu and Yang}]{21}
\bibinfo{author}{Li, Y.}, \bibinfo{author}{Fang, C.}, \bibinfo{author}{Yang,
  J.}, \bibinfo{author}{Wang, Z.}, \bibinfo{author}{Lu, X.},
  \bibinfo{author}{Yang, M.H.}, \bibinfo{year}{2017}b.
\newblock \bibinfo{title}{Universal style transfer via feature transforms}.
\newblock \bibinfo{journal}{Advances in neural information processing systems}
  \bibinfo{volume}{30}.
%Type = Article
\bibitem[{Li et~al.(2017c)Li, Wang, Liu and Hou}]{demystifying}
\bibinfo{author}{Li, Y.}, \bibinfo{author}{Wang, N.}, \bibinfo{author}{Liu,
  J.}, \bibinfo{author}{Hou, X.}, \bibinfo{year}{2017}c.
\newblock \bibinfo{title}{Demystifying neural style transfer}.
\newblock \bibinfo{journal}{arXiv preprint arXiv:1701.01036} .
%Type = Inproceedings
\bibitem[{Lin et~al.(2014)Lin, Maire, Belongie, Hays, Perona, Ramanan,
  Doll{\'a}r and Zitnick}]{coco}
\bibinfo{author}{Lin, T.Y.}, \bibinfo{author}{Maire, M.},
  \bibinfo{author}{Belongie, S.}, \bibinfo{author}{Hays, J.},
  \bibinfo{author}{Perona, P.}, \bibinfo{author}{Ramanan, D.},
  \bibinfo{author}{Doll{\'a}r, P.}, \bibinfo{author}{Zitnick, C.L.},
  \bibinfo{year}{2014}.
\newblock \bibinfo{title}{Microsoft coco: Common objects in context}, in:
  \bibinfo{booktitle}{Computer Vision--ECCV 2014: 13th European Conference,
  Zurich, Switzerland, September 6-12, 2014, Proceedings, Part V 13},
  \bibinfo{organization}{Springer}. pp. \bibinfo{pages}{740--755}.
%Type = Inproceedings
\bibitem[{Luo et~al.(2022)Luo, Han and Yang}]{pama}
\bibinfo{author}{Luo, X.}, \bibinfo{author}{Han, Z.}, \bibinfo{author}{Yang,
  L.}, \bibinfo{year}{2022}.
\newblock \bibinfo{title}{Progressive attentional manifold alignment for
  arbitrary style transfer}, in: \bibinfo{booktitle}{Proceedings of the Asian
  Conference on Computer Vision}, pp. \bibinfo{pages}{3206--3222}.
%Type = Inproceedings
\bibitem[{Park and Lee(2019)}]{SANet}
\bibinfo{author}{Park, D.Y.}, \bibinfo{author}{Lee, K.H.},
  \bibinfo{year}{2019}.
\newblock \bibinfo{title}{Arbitrary style transfer with style-attentional
  networks}, in: \bibinfo{booktitle}{proceedings of the IEEE/CVF conference on
  computer vision and pattern recognition}, pp. \bibinfo{pages}{5880--5888}.
%Type = Article
\bibitem[{Phillips and Mackintosh(2011)}]{wiki}
\bibinfo{author}{Phillips, F.}, \bibinfo{author}{Mackintosh, B.},
  \bibinfo{year}{2011}.
\newblock \bibinfo{title}{Wiki art gallery, inc.: A case for critical
  thinking}.
\newblock \bibinfo{journal}{Issues in Accounting Education}
  \bibinfo{volume}{26}, \bibinfo{pages}{593--608}.
%Type = Article
\bibitem[{Risser et~al.(2017)Risser, Wilmot and Barnes}]{27}
\bibinfo{author}{Risser, E.}, \bibinfo{author}{Wilmot, P.},
  \bibinfo{author}{Barnes, C.}, \bibinfo{year}{2017}.
\newblock \bibinfo{title}{Stable and controllable neural texture synthesis and
  style transfer using histogram losses}.
\newblock \bibinfo{journal}{arXiv preprint arXiv:1701.08893} .
%Type = Article
\bibitem[{Sengupta et~al.(2019)Sengupta, Ye, Wang, Liu and Roy}]{vgg}
\bibinfo{author}{Sengupta, A.}, \bibinfo{author}{Ye, Y.},
  \bibinfo{author}{Wang, R.}, \bibinfo{author}{Liu, C.}, \bibinfo{author}{Roy,
  K.}, \bibinfo{year}{2019}.
\newblock \bibinfo{title}{Going deeper in spiking neural networks: Vgg and
  residual architectures}.
\newblock \bibinfo{journal}{Frontiers in neuroscience} \bibinfo{volume}{13},
  \bibinfo{pages}{95}.
%Type = Inproceedings
\bibitem[{Shen et~al.(2018)Shen, Yan and Zeng}]{meta}
\bibinfo{author}{Shen, F.}, \bibinfo{author}{Yan, S.}, \bibinfo{author}{Zeng,
  G.}, \bibinfo{year}{2018}.
\newblock \bibinfo{title}{Neural style transfer via meta networks}, in:
  \bibinfo{booktitle}{Proceedings of the IEEE Conference on Computer Vision and
  Pattern Recognition}, pp. \bibinfo{pages}{8061--8069}.
%Type = Article
\bibitem[{ShiQi~Jiang(2023a)}]{CTDP}
\bibinfo{author}{ShiQi~Jiang, JunJie~Kang, Y.L.}, \bibinfo{year}{2023}a.
\newblock \bibinfo{title}{Color and texture dual pipeline lightweight style
  transfer}.
\newblock \bibinfo{journal}{arXiv preprint arXiv:2310.01321} .
%Type = Article
\bibitem[{ShiQi~Jiang(2023b)}]{DcDae}
\bibinfo{author}{ShiQi~Jiang, JunJie~Kang, Y.L.}, \bibinfo{year}{2023}b.
\newblock \bibinfo{title}{Degree-controllable lightweight fast style transfer
  with detail attention-enhanced}.
\newblock \bibinfo{journal}{arXiv preprint arXiv:2306.16846} .
%Type = Article
\bibitem[{ShiQi~Jiang(2023c)}]{IDD}
\bibinfo{author}{ShiQi~Jiang, JunJie~Kang, Y.L.}, \bibinfo{year}{2023}c.
\newblock \bibinfo{title}{Dual pipeline style transfer with input distribution
  differentiation}.
\newblock \bibinfo{journal}{arXiv preprint arXiv:2311.05432} .
%Type = Article
\bibitem[{Ulyanov et~al.(2016a)Ulyanov, Lebedev, Vedaldi and Lempitsky}]{32}
\bibinfo{author}{Ulyanov, D.}, \bibinfo{author}{Lebedev, V.},
  \bibinfo{author}{Vedaldi, A.}, \bibinfo{author}{Lempitsky, V.},
  \bibinfo{year}{2016}a.
\newblock \bibinfo{title}{Texture networks: Feed-forward synthesis of textures
  and stylized images}.
\newblock \bibinfo{journal}{arXiv preprint arXiv:1603.03417} .
%Type = Article
\bibitem[{Ulyanov et~al.(2016b)Ulyanov, Vedaldi and Lempitsky}]{33}
\bibinfo{author}{Ulyanov, D.}, \bibinfo{author}{Vedaldi, A.},
  \bibinfo{author}{Lempitsky, V.}, \bibinfo{year}{2016}b.
\newblock \bibinfo{title}{Instance normalization: The missing ingredient for
  fast stylization}.
\newblock \bibinfo{journal}{arXiv preprint arXiv:1607.08022} .
%Type = Inproceedings
\bibitem[{Wang et~al.(2020)Wang, Li, Wang, Hu and Yang}]{collaborative}
\bibinfo{author}{Wang, H.}, \bibinfo{author}{Li, Y.}, \bibinfo{author}{Wang,
  Y.}, \bibinfo{author}{Hu, H.}, \bibinfo{author}{Yang, M.H.},
  \bibinfo{year}{2020}.
\newblock \bibinfo{title}{Collaborative distillation for ultra-resolution
  universal style transfer}, in: \bibinfo{booktitle}{Proceedings of the
  IEEE/CVF conference on computer vision and pattern recognition}, pp.
  \bibinfo{pages}{1860--1869}.
%Type = Inproceedings
\bibitem[{Wang et~al.(2017)Wang, Oxholm, Zhang and Wang}]{35}
\bibinfo{author}{Wang, X.}, \bibinfo{author}{Oxholm, G.},
  \bibinfo{author}{Zhang, D.}, \bibinfo{author}{Wang, Y.F.},
  \bibinfo{year}{2017}.
\newblock \bibinfo{title}{Multimodal transfer: A hierarchical deep
  convolutional neural network for fast artistic style transfer}, in:
  \bibinfo{booktitle}{Proceedings of the IEEE conference on computer vision and
  pattern recognition}, pp. \bibinfo{pages}{5239--5247}.
%Type = Article
\bibitem[{Wang et~al.(2021)Wang, Zhao, Chen, Zuo, Li, Xing and Lu}]{wang}
\bibinfo{author}{Wang, Z.}, \bibinfo{author}{Zhao, L.}, \bibinfo{author}{Chen,
  H.}, \bibinfo{author}{Zuo, Z.}, \bibinfo{author}{Li, A.},
  \bibinfo{author}{Xing, W.}, \bibinfo{author}{Lu, D.}, \bibinfo{year}{2021}.
\newblock \bibinfo{title}{Divswapper: towards diversified patch-based arbitrary
  style transfer}.
\newblock \bibinfo{journal}{arXiv preprint arXiv:2101.06381} .
%Type = Article
\bibitem[{Wang et~al.(2022)Wang, Zhao, Zuo, Li, Chen, Xing and Lu}]{microast}
\bibinfo{author}{Wang, Z.}, \bibinfo{author}{Zhao, L.}, \bibinfo{author}{Zuo,
  Z.}, \bibinfo{author}{Li, A.}, \bibinfo{author}{Chen, H.},
  \bibinfo{author}{Xing, W.}, \bibinfo{author}{Lu, D.}, \bibinfo{year}{2022}.
\newblock \bibinfo{title}{Microast: Towards super-fast ultra-resolution
  arbitrary style transfer}.
\newblock \bibinfo{journal}{arXiv preprint arXiv:2211.15313} .
%Type = Inproceedings
\bibitem[{Xie et~al.(2022)Xie, Li, Huang, Fu, Wang and Guo}]{xie}
\bibinfo{author}{Xie, X.}, \bibinfo{author}{Li, Y.}, \bibinfo{author}{Huang,
  H.}, \bibinfo{author}{Fu, H.}, \bibinfo{author}{Wang, W.},
  \bibinfo{author}{Guo, Y.}, \bibinfo{year}{2022}.
\newblock \bibinfo{title}{Artistic style discovery with independent
  components}, in: \bibinfo{booktitle}{Proceedings of the IEEE/CVF Conference
  on Computer Vision and Pattern Recognition}, pp.
  \bibinfo{pages}{19870--19879}.
%Type = Inproceedings
\bibitem[{Zhang et~al.(2019)Zhang, Zhu and Zhu}]{zhang}
\bibinfo{author}{Zhang, C.}, \bibinfo{author}{Zhu, Y.}, \bibinfo{author}{Zhu,
  S.C.}, \bibinfo{year}{2019}.
\newblock \bibinfo{title}{Metastyle: Three-way trade-off among speed,
  flexibility, and quality in neural style transfer}, in:
  \bibinfo{booktitle}{Proceedings of the AAAI Conference on Artificial
  Intelligence}, pp. \bibinfo{pages}{1254--1261}.
%Type = Inproceedings
\bibitem[{Zhang and Dana(2018a)}]{37}
\bibinfo{author}{Zhang, H.}, \bibinfo{author}{Dana, K.}, \bibinfo{year}{2018}a.
\newblock \bibinfo{title}{Multi-style generative network for real-time
  transfer}, in: \bibinfo{booktitle}{Proceedings of the European Conference on
  Computer Vision (ECCV) Workshops}, pp. \bibinfo{pages}{0--0}.
%Type = Inproceedings
\bibitem[{Zhang and Dana(2018b)}]{12}
\bibinfo{author}{Zhang, H.}, \bibinfo{author}{Dana, K.}, \bibinfo{year}{2018}b.
\newblock \bibinfo{title}{Multi-style generative network for real-time
  transfer}, in: \bibinfo{booktitle}{Proceedings of the European Conference on
  Computer Vision (ECCV) Workshops}, pp. \bibinfo{pages}{0--0}.

\end{thebibliography}

\end{document}